# Pedestrian Models for Autonomous Driving Part II: High-Level Models of Human Behavior

Fanta Camara[1,2], Nicola Bellotto[2], Serhan Cosar[3], Florian Weber[4], Dimitris Nathanael[5], Matthias Althoff[6], Jingyuan Wu[7], Johannes Ruenz[7], André Dietrich[8], Gustav Markkula[1], Anna Schieben[9], Fabio Tango[10], Natasha Merat[1] and Charles Fox[1,2,11]

*Abstract*—Autonomous vehicles (AVs) must share space with pedestrians, both in carriageway cases such as cars at pedestrian crossings and off-carriageway cases such as delivery vehicles navigating through crowds on pedestrianized high-streets. Unlike static obstacles, pedestrians are active agents with complex, interactive motions. Planning AV actions in the presence of pedestrians thus requires modelling of their probable future behaviour as well as detecting and tracking them. This narrative review article is Part II of a pair, together surveying the current technology stack involved in this process, organising recent research into a hierarchical taxonomy ranging from low-level image detection to high-level psychological models, from the perspective of an AV designer. This self-contained Part II covers the higher levels of this stack, consisting of models of pedestrian behaviour, from prediction of individual pedestrians' likely destinations and paths, to game-theoretic models of interactions between pedestrians and autonomous vehicles. This survey clearly shows that, although there are good models for optimal walking behaviour, high-level psychological and social modelling of pedestrian behaviour still remains an open research question that requires many conceptual issues to be clarified. Early work has been done on descriptive and qualitative models of behaviour, but much work is still needed to translate them into quantitative algorithms for practical AV control.

## I. INTRODUCTION

To operate successfully in the presence of pedestrians, autonomous vehicles require input from a huge variety of models that have to work seamlessly together. These models range from simple visual models for detection of pedestrians, to predicting their future movements using psychological and sociological methods. Part I of this two-part survey [33] covered models for sensing, detection, recognition, and tracking

This project has received funding from EU H2020 interACT (723395).
[1] Institute for Transport Studies (ITS), University of Leeds, UK
[2] Lincoln Centre for Autonomous Systems, University of Lincoln, UK
[3] Institute of Engineering Sciences, De Montfort University, UK
[4] Bayerische Motoren Werke Aktiengesellschaft (BMW), Germany
[5] School of Mechanical Engineering, Nat. Tech. University of Athens
[6] Department of Computer Science, Technische Universität München
[7] Robert Bosch GmbH, Germany
[8] Chair of Ergonomics, Technische Universität München (TUM), Germany
[9] DLR (German Aerospace Center), Germany
[10] Centro Ricerche Fiat (CRF), Italy
[11] Ibex Automation Ltd, UK
Manuscript received 2019-03-11; Revisions: 2019-10-21, 2020-03-26. Accepted: 09-04-2020.

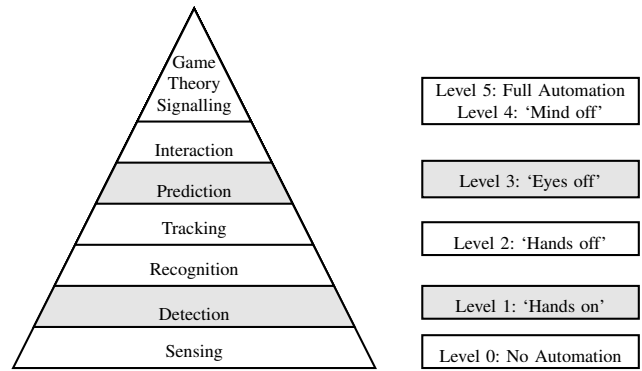

Fig. 1. Main structure of the review.

of pedestrians. Part II here reviews models for pedestrian trajectory prediction, interaction of pedestrians, and behavioral modelling of pedestrians, and also experimental resources to validate all the types of models. Interacting with pedestrians is a particular type of social intelligence. Autonomous vehicles will need to utilize many different levels of models of pedestrians, each addressing different aspects of perception and action. Each of these models can be based on empirical science results or obtained via machine learning. In contrast to the models of Part I, Part II requires models from higher levels of the technology stack, as researched by psychologists and taught in advanced driver training programmes. For instance, drivers often try to infer the personality of other humans, predict their likely behaviours, and interact with them to communicate mutual intentions [102]. Between the high level surveyed in this Part II and the low levels of Part I, researchers infer psychological information from perceptual information. As an example, researchers build systems to recognize the body language, gestures, and demographics information of pedestrians to better predict their likely goals and behaviours. Despite the importance of bridging the research between the higher and lower levels, their connection is still thin, both conceptually and in terms of actual implementations.

While prediction of likely future pedestrian trajectories is becoming increasingly well understood, models for actively controlling pedestrian interactions – including game-theoretic models – are still in their infancy. Active control here means that the vehicle's own future actions are taken into account in predicting how the pedestrian will respond, and vice versa. One reason is that sufficient data to rigorously study interac-

TABLE I
PROPOSED MAPPING FROM SAE LEVELS TO PEDESTRIAN MODEL REQUIREMENTS.

| SAE LEVEL | DESCRIPTION | MODEL REQUIREMENTS | SECTION |
|---|---|---|---|
| 0 | No Automation. Automated system issues warnings and may momentarily intervene, but has no sustained vehicle control. | Sensing | Part I [33] Sec. II |
| 1 | Hands on. The driver and the automated system share control of the vehicle. For example, adaptive cruise control (ACC), where the driver controls steering and the automated system controls speed. The driver must be ready to resume full control when needed. | +Detection | Part I [33] Sec. III |
| 2 | Hands off. The automated system takes full control of the vehicle (steering and speed). The driver must monitor and be prepared to intervene immediately. Occasional contact between hand and wheel is often mandatory to confirm that the driver is ready to intervene. | +Recognition<br>+Tracking | Part I [33] Sec. IV<br>Part I [33] Sec. V |
| 3 | Eyes off. Driver can safely turn attention away from the driving tasks, e.g. use a phone or watch a movie. Vehicle will handle situations that call for an immediate response, like emergency braking. The driver must still be prepared to intervene within some limited time. | +Unobstructed Walking Models, Known Goals<br>+Behaviour Prediction, Known Goals<br>+Behaviour Prediction, Unknown Goals | Sec. II-A<br>Sec. II-B<br>Sec. II-C |
| 4 | Mind off. No driver attention is required for safety, except in limited spatial areas (geofenced) or under special circumstances, like traffic jams. Outside of these areas or circumstances, the vehicle must be able to safely abort or transfer control to the human. | +Event/Activity Models<br>+Effects of Pedestrian Class on Trajectory<br>+Pedestrian Interaction Models<br>+Game Theoretic and Signalling Models | Sec. II-D<br>Sec. II-E<br>Sec. III<br>Sec. IV |
| 5 | Full automation. No human intervention is required at all, fully automated driving. | +Extreme Robustness and Reliability | |

Note: '+X' means that 'X' is required in addition to the requirements of the previous level.

tion between pedestrians has only recently become available as presented in Sec. V on experimental resources. Another reason is that one first has to be able to reliably sense, detect, recognize, and track pedestrians in order to gather enough data for modelling interaction and game-theoretic models. A third reason is that interaction and game-theoretic models are only relevant in crowded environments, while many situations do not require much interaction. However, crowded environments are those that are typically most relevant for autonomous driving. Fig. 1 shows the review structure.

To assess the maturity of the methods presented, the level of autonomy is used, as defined by the Society of Automotive Engineers (SAE) – the same measure has already been used in Part I [33]. For the convenience of the reader, the five SAE levels are briefly presented, ranging from simple driver assistance tools to full self-driving [183]. Requirements for pedestrian modelling increase with each level, with lower levels typically requiring lower and more mature levels of pedestrian models, such as detection and tracking, while higher levels require models for psychological and social understanding to fully interact with pedestrians in a human-like way [30]. Table I gives an overview of SAE levels and requirements mappings.

While many papers propose pedestrian models at various levels, no unifying theory has yet been produced which would make it possible to easily transfer results across all levels from detection to prediction. This review uncovers bottlenecks in transferring results to facilitate closing existing research gaps. Also, many existing studies only consider results from empirical science or those obtained via machine learning. This survey provides an overview considering both possibilities. While machine learning results work particularly well for detection and recognition, they are not yet performing so well for prediction. Some reasons are that prediction is a more high-dimensional problem, with dimensions including goals, obstacles, various state variables of pedestrians, and road geometry. A further reason is that less labelled data is available for training prediction models. A promising future direction is to combine empirical science results with machine learning to better safeguard techniques using machine learning and to avoid over-fitting.

While similar concepts apply to modelling human drivers and their vehicles for interactions with AVs, this article presents a review of the state of the art specifically in modelling human pedestrians for social decision-making. In some cases it goes beyond modelling aspects to also cover more conceptual aspects or empirical psychological findings, when the studies in question are judged to have very direct applicability to mathematical models. Results from human driving cannot be directly translated to pedestrians due to the variability in locomotion, the differences in shape, the changes in postures and the less-structured environment.

*Pedestrians* are defined as humans moving on and near public highways including roads and pedestrianised areas, who walk using their own locomotive power. This excludes, for example, humans moving on cycles, wheelchairs and other mobility devices, skates and skateboards, or those transported by other humans. This review does not cover interactions of traffic participants without pedestrians: a survey on trajectory prediction of on-road vehicles is provided in [123] and a survey on vision-based trajectory learning is provided in [146].

This Part II is organized as shown in Fig. 2. In Sec. II, methods for predicting the movements of pedestrians are reviewed. In particular, we consider models and methods for unstructured environments, for prediction around obstacles, to estimate destinations, and for the prediction of events such as crossing the road. These methods are enhanced in Sec. III for groups of pedestrians interacting with each other. This section considers the complete variety of researched models from macroscopic models only considering flow of people to microscopic models that consider individual pedestrians. In many situations, interaction models do not require game theory, because pedestrians often have different goals. However, there are also many situations, where pedestrians have competing goals, e.g., when several pedestrians have to pass a narrow passage. In such situations, the game theoretic models presented in Sec. IV can be very useful. Finally, Sec. V surveys available resources: datasets and simulators, both for pedestrians and vehicles.

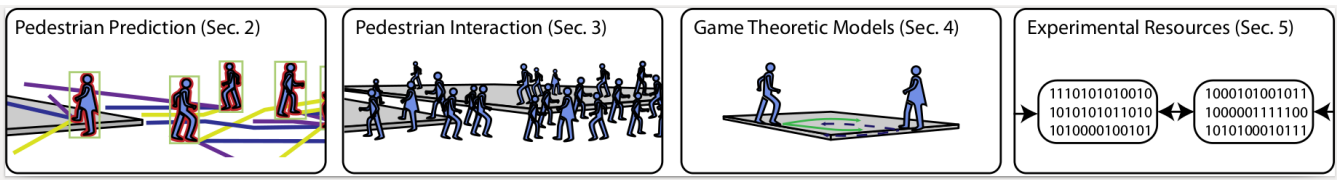

Fig. 2. Structure of the paper.

## II. BEHAVIOUR MODELS WITHOUT INTERACTION

The tracking models reviewed in Part I are *kinematic* in that they assume that pedestrians move in physical and/or pose space in motion described by kinematic models. This is a very basic assumption – human drivers typically have much more complex understandings and hence predictions of pedestrian behavior which they use to drive safely in their presence [102]. These range from slightly more advanced kinematic understandings such as 'pedestrians tend to walk in straight lines' to models of how they are likely to interact with static objects in their environment, and predictions of pedestrians' likely destinations from reading the street scene.

This section reviews such models starting from simple unobstructed path models to uncertain destination models and more advanced event/activity models. These models do not yet consider interaction with other agents. Figure 3 summarizes the classes of models presented in this section. A previous review was proposed by Ridel *et al.* [172], which mainly considered pedestrian crossing intent and offered a restricted view of the different models developed for trajectory prediction.

### A. Unobstructed Walking Models with Known Goals

Given a start location and orientation, and a goal location, humans do not typically turn towards the goal on the spot (which would waste time) and then walk in a straight line, but rather set off walking in their initial heading and adjust their orientation gradually as they walk, resulting in smooth, curved trajectories from origin to destination [72]. Models from optimal control theory as also used in robotics [50] define cost functions for travel time, speed, and accelerations, to reproduce these characteristic curved trajectories. The model in [72] instead achieves curved trajectories by modelling the rate of turning of the pedestrian as a function of the visual angle and distance to the goal. A simple kinematic model consists in considering human locomotion as a nonholonomic motion [161], using the unicycle model (1) where the pedestrian walking trajectory is represented by the trajectory of their center of gravity, 2D coordinates $(x, y)$ and by the angle $\theta$,

$$\begin{aligned}\dot{x} &= u_1 \cos\theta \\ \dot{y} &= u_1 \sin\theta \\ \dot{\theta} &= u_2\end{aligned} \quad (1)$$

where $u_1$ is the forward velocity and $u_2$ is the angular velocity. Assuming known origin and destination with inverse optimal control, one can reliably predict human walking paths using this model [9] [155].

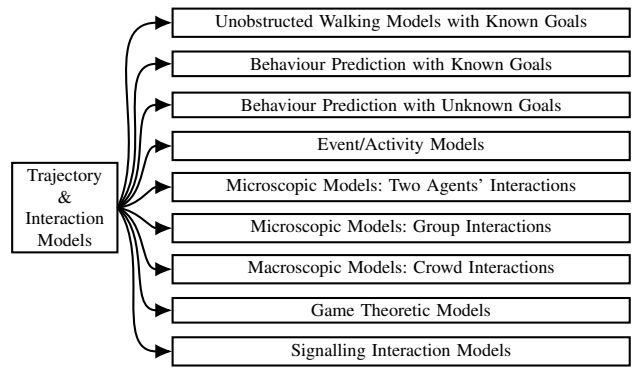

Fig. 3. Pedestrian behaviour prediction and interaction models.

### B. Behaviour Prediction with Known Goals

Here, the likely behaviour of a pedestrian in a static environment is considered, given a map. Pedestrians are likely to route around obstacles, and to stop at the edges of roads before crossing. This section does not consider social effects of other agents – this is presented later in Sec. III.

*1) Dynamic Graphical Models:* Dynamic Graphical Models (DGM) are Graphical Models of a particular topology, containing some Markovian sequence of variables over time. DGMs include simple Markov and Hidden Markov Models and also more complex models. The method in [145] used tracking in a DGM based on particle filter approximation to infer beliefs over future pedestrian trajectories and combined this with a GNSS (Global Navigation Satellite System) module that provides information about the hazardous areas and people.

*2) Gaussian Process Methods:* Habibi *et al.* [88] proposed a context-based approach to pedestrian trajectory prediction using Gaussian Processes [166]. This model incorporates context features such as the pedestrian's distance to the traffic light, the distance to the curbside, and the curbside orientation in the transition learning phase to improve the prediction. A context-based augmented semi non-negative sparse coding (CASNSC) algorithm is used to predict pedestrian trajectories.

*3) Deep Learning Methods:* Bock *et al.* [24] developed a Recurrent Neural Network (LSTM) model to learn pedestrian behaviour patterns at intelligent intersections using camera data from the onboard vehicle and the infrastructure. The model can predict trajectories for a horizon of 5s.

*4) Other Methods:* Kruse *et al.* [121] was one of the first attempts to statistically infer human motion patterns from data and incorporate them in a robot motion planner for obstacle avoidance. Garzón *et al.* [77] presented a pedestrian trajectory prediction model based on two path planning algorithms that require a set of possible goals, a map and the initial position. It then computes similarities between the obtained and observed trajectories into probabilities. This model is run along with a pedestrian detector and tracker. Tamura *et al.* [198] proposed a pedestrian behaviour model that is based on social forces and takes into account the intention of the pedestrian in the trajectory prediction by defining a set of subgoals. In [170] the uncertain goals are used as latent variables to guide the motion prediction of pedestrians. Their positions are predicted by combining forward propagation of a physical model with local a priori information (e.g., obstacles and different road types) from the start position, and by planning the trajectory from a goal position. The distribution over the destinations is modeled with a particle filter.

In [209], Vasishta *et al.* presented a model based on the principle of natural vision that incorporates contextual information extracted from the environment to the pedestrian behavior and it especially tries to predict hazardous behavior such as crossing in non authorized areas. The aforementioned model in [72] considers goals and obstacles as distance-dependent attractors and repellers in heading angle space. The contributions from the goal and obstacles are linearly combined, yielding a momentary rate of acceleration of heading, which results in human-like trajectories for simultaneous goal-seeking and obstacle avoidance. In [57], Dias *et al.* developed a model simulating pedestrian behaviour around corners, using minimum jerk theory and one-thirds power law concept. Their model uses Monte Carlo simulation to generate pedestrian trajectories with turning maneuvers, which were comparable to empirical trajectories.

*C. Behaviour Prediction with Unknown Goals*

Many of the above models assume known probable destinations for pedestrians, which enable routing to act not just around local obstacles, but to predict entire long-term trajectories, such as for pedestrians intending to cross the road. However, in reality a pedestrian's destination is rarely given.

*1) Dynamic Graphical Models:* Ziebart *et al.* [233] presented a pedestrian trajectory prediction model that takes into account hindrance due to robot motion, as is required in off-carriageway interactions such as last mile AVs in pedestrianized areas. A maximum entropy inverse optimal control technique, introduced in [232], is used and is equivalent to a soft-maximum version of Markov decision process (MDP) that accounts for decision uncertainty into the trajectories distribution. The cost function is a linear combination of the features (e.g obstacles) in the environment. People's motion can be modeled by an MDP and by choosing a certain path, there is an immediate reward. The model is conditioned on a known destination location but the model reasons about all possible destinations and the real destination is not known at the prediction time. The destination is inferred in a Bayesian way, by computing the prior distributions over destinations using previous observed trajectories. When there is no previous data, features (door, chair etc.) in the environment are used to model the destination. In [113], Kitani *et al.* extended [232], [233] by incorporating visual features to forecast future activities and destinations. The observations provided by the vision system (e.g. tracking algorithm) are assumed to be noisy and uncertain therefore they used a hidden variable Markov decision process (hMDP) where the agent knows its own states, action and reward but observes only noisy measurements. Negative Log-Loss (NLL) is used as a probabilistic metric and Modified Hausdorff Distance (MHD) as a physical measure of the distance between two trajectories. Vasquez [210] extends the work of Ziebart [233] and Kitani [113] while reducing computational costs.

Bennewitz *et al.* [18], [17] proposed a learning method for human motion recognition using the expectation maximization (EM) and a hidden Markov model (HMM) for clustering and predicting human trajectories and incorporating them into a robot path planner. In [221], Wu *et al.* presented a model that uses Markov chains for pedestrian motion prediction (able to deal with non-Gaussian distribution and several constraints). A heuristic method is proposed to automatically infer the positions of several potential goals on a generic semantic map. It also incorporates policies to predict the pedestrian motion direction and takes into account other traffic participants by incorporating a collision checking approach. Borgers *et al.* [29] presented a model that predicts pedestrians' route choice based on Markov chains. Similarly, Bai *et al.* [11] presented a real-time approximate POMDP (Partially Observable Markov Decision Process) controller, DESPOT, for use in high-street type environments. The method is intention-aware in the sense of inferring pedestrian destinations and route plans from their observed motion over time, and accounting for the value of this information against the value of making progress while planning a robot's own route around them. Karasev *et al.* [110] presented a long-term prediction model that incorporates environmental constraints with the intent modeled by a policy in a MDP framework. The pedestrian state is estimated using a Rao-Blackwellized filter and pedestrian intent by planning according to a stochastic policy. This model assumes that pedestrians behave rationally.

*2) Deep Learning Methods:* Hug *et al.* [98] proposed a LSTM-MDL model combined with a particle filter method for multi-modal trajectory prediction, and tested on Stanford Drone Dataset (SDD) [176]. Rehder *et al.* [171] proposed a method to infer pedestrian destinations. The trajectory prediction is computed as a goal-oriented motion planning.

The whole system is based on deep-learning and trained via inverse reinforcement learning. A general introduction on reinforcement learning in robotics can be found in [115]. Deo et al. [56] presented a framework for multi-modal pedestrian trajectory forecasting in structured environments. They used a convolutional neural network to compute both the reward maps of the path states and the possible goal states for MDPs. The derived policy information is then fed into a recurrent neural network, combined with track history, to generate possible future trajectories. Goldhammer et al. [81] developed a Multilayer Perceptron (MLP) neural network with polynomial least square approximation to predict pedestrian trajectories based on camera data. A long-term prediction model using RNNs is proposed in [22].

*3) Other Methods:* Cosgun et al. [52] presented a person-following service robot with a task dependent motion planner. The robot can track and predict the future trajectory of the person by maximizing its reward at future steps while avoiding entering into the human's personal space. Koschi et al. [117] proposed a set-based method to predict all possible behaviours of pedestrians using reachability analysis [5] for pedestrian occupancy. Pedestrians are described as point mass with a certain maximum velocity and maximum acceleration. A rule-based occupancy is applied that does not allow a pedestrian to obstruct traffic, e.g. pedestrians are given priority at crosswalks and their trajectory is assumed to be evasive.

### D. Event/Activity Models

Pedestrian event models consider stereotypical sequences of behaviours of individual pedestrians. These may give additional information about route choice, beyond that available from static classification of the pedestrian. For example, a commuter, or class of commuters, who engage in similar actions every day, such as road crossing in a certain way then checking their phone, may reveal information about their identity to enable re-identification[1] which is in turn predictive of their future destinations. These models look for features predictive of route choice in static environments and do not consider social factors.

*1) Dynamic Graphical Methods:* Duckworth et al. [65] [64] developed on a mobile robot an unsupervised qualitative spatio-temporal relations (QSR) model to learn motion patterns using a graph representation and is able to predict people's future behaviour. Dondrup et al. [63] presented a ROS-based real-time human perception framework for mobile robots using laser and RGB-D data and tracking people with a Kalman filter approach. Human trajectories are converted into QSR (Qualitative Spatial Relations) and used for a Hidden Markov Model (HMM) to classify the behaviour of the different people encountered [62]. In [187], Schneider and Gavrila presented a comparative study on Bayesian filters (EKF and IMM) for short-term (<2s) pedestrian trajectory prediction, in particular they used stereo camera images to apply these methods to four different types of behaviour: crossing, stopping, bending in and starting.

---
[1]Identity here is distinct from personal information as defined by privacy laws such as the EU General Data Protection Regulation (GDPR).

Body heading is used above in basic path planning models, but head-turning events are distinct from body heading, and are discrete events which occur when a pedestrian turns their head to look around rather than to orient their body. Such an activity model is used in [116] to enhance path prediction of pedestrians while intending to cross a street. For low-level occupancy prediction, a dynamic Bayesian network (DFBN) is used on top of a switching linear dynamic system (SLDS) anticipating the changes of pedestrian dynamics. As in [116], studies [189], [190] also model head orientation by an event/activity model to enhance the underlying prediction approach.

*2) Gaussian Process Methods:* Quintero et al. [162] [163] proposed a pedestrian path prediction method up to 1s ahead based on balanced Gaussian Process dynamical models (B-GPDMs) and naïve Bayes classifiers. GPDM is used to transform a sequence of timed feature vectors into a low dimensional latent space and it can predict the next position based on the current one. The naïve Bayes classifiers are used to classify pedestrian actions based on 3D joint positions.

*3) Feature Selection Methods:* Bonnin et al. [27] proposed a generic context-based model to predict pedestrians behavior according to features describing their local urban environment. To learn about interactions between autonomous vehicles and pedestrian interactions, in [39], Camara et al. collected data from real-world pedestrian-vehicle interactions at an unsignalized intersection. The actions of pedestrians and vehicles were ordered into sequences of events comprising descriptive features and the study revealed the most predictive features in a crossing scenario such as the head direction, the position on the pavement, hand gestures etc. In [38], these features were filtered over time to predict whether the pedestrian would first cross the intersection or not. Völz et al. [213] [214] proposed a model that can predict whether or not a pedestrian will cross the street with a set of features learnt from a database of LIDAR pedestrian trajectories that are used as inputs for a support vector machine (SVM).

### E. Effects of Pedestrian Class on Trajectory

The models reviewed so far consider all pedestrians to be alike, but human drivers interacting with pedestrians may consider their attributes as members of stereotypical classes. Membership of various demographic and psychological state classes may be predictive of their behaviour. This section first reviews findings from the psychological literature suggesting what such classes could be usefully predictive of behaviour, if it was possible to classify them automatically from autonomous vehicles. Rasouli and Tsotsos reviewed pedestrian demographics for interactions with autonomous vehicles and argued that knowing such information could help AVs, cf. Sec. III. 1. in [169]. Figure 4 presents a set of pedestrian attributes used for behaviour modelling.

*a) Effects of Age and Gender:* Wilson et al. [219] performed a large-scale study on adult pedestrian crossing behavior and concluded that elderly people take more time and have more head movements during the crossing. Evans et al. [71] used the Theory of Planned Behavior (TPB) [1] via a questionnaire to predict adolescents' intentions during a hazardous road-crossing scenario. Their results show that older and male adolescents had stronger intentions to cross and that moral norms do not have any influence on crossing decisions. Pedestrians who considered themselves as safe pedestrians were less likely to cross and the anticipated affective reactions were important. Bernhoft and Carstensen [21] compared the crossing preferences and behaviour of elderly pedestrians and cyclists (age 70+) to younger people aged 40-49. It was found that elderly people have a preference for road facilities that they consider to be safer such as pavements, pedestrian crossings, signalized intersections, cycle paths. The differences between the two groups are said to be related to health and physical abilities of the people rather than their differences in age and gender.

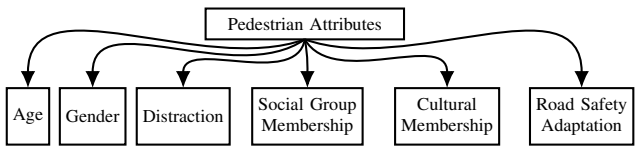

Fig. 4. Pedestrian behaviour attributes.

Several studies have shown that older pedestrians have a larger accident rate than younger people [219]. Gorrini et al. [83] also found differences in adults and elderly people crossing behaviour. The study of Oxley *et al* [153] showed that older pedestrians have more risky crossing behavior in complex traffic environments than younger people. Not surprisingly, many authors have found decreasing crossing speeds with age [10] [130], [202], compensated for by requiring larger time gaps in traffic before commencing crossing [130]. In addition, Avineri *et al.* [10] found lower crossing speeds for female than male pedestrians, and that the fear of falling in elderly pedestrians has an effect on the number of downward head pitches during crossing. Holland and Hill [95] used the TPB for pedestrians' intention analysis while crossing the road. The results showed that women perceived more risk and were less likely to cross than men. In [96], they also studied the effect of gender on pedestrian crossing behaviour and showed that men with a driving experience make safer crossings than non-drivers and that older women were found to make more unsafe crossing decisions than younger women.

*b) Distraction:* Distraction of pedestrians from traffic environments would ideally be defined via their mental state i.e., thinking about a problem unrelated to their environment; or approximated in practice via observable proxies. While it is possible that mental distraction might be measurable via hard-to-observe proxies such as gaze direction or high-level body language, it may be more practical to look instead for known *causes* of distraction. Schwebel *et al.* [191] performed a study in a semi-immersive virtual pedestrian street with college students, finding an impact of talking on mobile phones on crossing behaviour. Walker *et al.* [215] showed that male pedestrians using a personal music device were more cautious in crossing than those who were not distracted. In [200], the effects of personal electronic device usage on crossing behavior is studied. The results show a third of the observed pedestrians were distracted by their mobile phone and that distracted pedestrians are more likely to have unsafe crossing behaviour and walk much faster than undistracted pedestrians.

*c) Social Group Membership:* Group membership can affect road crossing. Three strangers in a group are less likely to assert in a crossing than three friends. In particular, group size influences a lot crossing behavior [169]. Zeedyk *et al.* [226] performed a study with adult-child pairs while crossing the road at a pedestrian crossing. They found that adults were more likely to hold girls' hands than boys.

*d) Cultural Membership:* In contrast to the above membership of short-term, physically present groups, it is also possible to consider 'cultural membership' of a pedestrian to any long-term, non-physically present group that may be usefully predictive of behaviour. For example, it might be possible for a human driver or autonomous vehicle to classify pedestrians as members of religious, sporting, or musical (sub)cultures as a probabilistic function of features of their clothing such as shape and colour of garments or symbols displayed on them; and that members of such groups show statistically significant differences in assertiveness, politeness, and other road interaction behaviours (cf. [169]). In Sociology, classifications of individuals into cultures is notoriously problematic and politicised. But for the purpose of predicting road interactions, *any* classification derived from observable features may be usefully considered if it improves predictions.

*e) Road Safety Adaptation:* Related to the possible predictiveness of cultural clothing is the effect of road safety clothing on behaviour. Human drivers are more likely to yield to pedestrians wearing high-visibility clothing [92], so it is also possible that knowing this fact will make a pedestrian wearing such clothing more likely to behave assertively. This is an example of risk compensation *adaptation*, a well-known effect in road safety in which the owners of safety improvements make economic decisions whether to use them to reduce accidents or alternatively to gain some other advantage at the cost of retaining the original accident rates [180].

*F. Discussion*

Single pedestrian unobstructed walking path and behaviour prediction around obstacles for known origins and destinations has well-established solutions. Their main strength lies in their simplicity and ease of implementation but their applicability to solve real AV problems is very limited due to the strong assumptions (e.g static obstacles, known origin-destination of pedestrians) which are not easily verified in the real world. But when – as is usual in real-time systems – the destinations of pedestrians are not known in advance, trajectory prediction is harder and remains an open research area.

Uncertain destination models may use known destination models as a subcomponent and average over them weighted by predictions about what the destination is. To predict what a pedestrian's destination will be, many medium and high-level sources of information may be relevant and useful, if suitable models can be found. These models split roughly into short-term models for prediction horizons around 1-2s and long-term models predicting for a horizon of around 5-6s. Event-based models of activity assume that behaviour often contains repeated stereotypical chunks of behavior, which once recognised in early stages can predict their later stages. The major emerging long-term prediction methods rely on neural network ('deep learning) methods. There is a need to verify how the data-driven methods such as [6] can be actually applied online for real-time systems. These models can help AVs to more accurately predict single pedestrian behaviour for shorter or longer time horizons, e.g. to know precisely whether a pedestrians trajectory would interfere with the AV's own path. But their main challenges lie in their computational cost, which increases significantly with the number of destination guesses, with longer time horizons and the amount of data needed for learning pedestrian motion patterns. Moreover, deep learning models are sometimes referred to as 'black-box models, in the sense that AI developers cannot fully explain some decisions (e.g. feature selection) made by the neural networks, rendering them potentially problematic for investigating the causes of incidents involving AVs and for determining their liabilities [43][84].

Single pedestrians destinations and behaviours may be informed by their class memberships, including their demographics and other visible features, such as clothing types. There are many recent sociological studies giving evidence of these effects, but they have not yet been translated into algorithms suitable for autonomous vehicle use, which would be a promising new research area. It is conjectured that additional information about pedestrians' emotion states would be similarly informative (e.g. angry pedestrians more likely to assert themselves in competitions for road space), but no studies were found in this area. Traditionally, emotional state has been difficult to capture and record, so that manually annotation of data sets are too small for machine learning to use. But as machine vision for face and body language recognition continues to improve (cf. Part I [33] Sect. IV), they are expected to produce big data sets which will enable machine learning to operate and inform destination and behaviour predictions.

## III. PEDESTRIAN INTERACTION MODELS

So far, only path prediction models for single pedestrians in static environments ignoring interactions with other pedestrians have been reviewed. This section will consider models of interaction between pedestrians. In Social Science, pedestrian behavior models have been studied for a long time: a survey is provided in [42] [201]. These models can be classified in two categories, namely microscopic models and macroscopic models, as reviewed in [211]. Microscopic models model only each pedestrian individually. Macroscopic models do not model individual pedestrians and instead model the behaviour of a single aggregate entity such as a crowd or a flow. Papadimitriou *et al.* [154] presented a review on pedestrian behavior models and a study on pedestrian and crowd dynamics was proposed by Vizzari and Bandini in [212]. Bellomo *et al.* [15] reviewed mathematical models of vehicular traffic and crowds while Duives *et al.* [66] surveyed pedestrian crowd

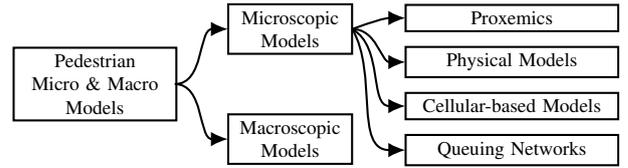

Fig. 5. Pedestrian microscopic and macroscopic models.

simulation models. Figure 5 presents a summary of pedestrian microscopic and macroscopic models.

### A. Microscopic Models

This section first describes pedestrian behaviour models at the microscopic level. It then presents pedestrian interaction models using these behaviour models for two agents' interactions and group behaviour modelling.

*1) Behaviour Models:* Microscopic models are divided into three main groups: physical models, cellular-based models and queuing network models. Each model is generally structured by two terms: one term that represents the attractive effects of pedestrians toward their goal and the other repulsive effects among and between pedestrians and the obstacles [42]. Proxemics is first described in this section.

*a) Proxemics:* The Psychology theory of Proxemics [91] studies human preferences (utilities) for having other humans in their proximity. Proxemics typically identifies four radial comfort zones, whose radii differ between cultures, for intimate, personal, social, and public space. These zones can be described by eight dimensions [91]:

1) postural-sex identifiers
2) sociofugal-sociopetal orientation (SFP axis)
3) kinesthetic factors
4) touch code
5) retinal combinations
6) thermal code
7) olfaction code
8) voice loudness scale

This model has been empirically tested with participants [217]. The theory is of great interest to pedestrian interaction models because it provides a possibly hard-wired negative utility not just for actual collisions with pedestrians but also for simply feeling too close to them. In particular, this provides a method for an AV to inflict a real negative utility on a pedestrian without touching them or risking their physical harm. Binary proxemics is the simplest case used in simple models, in which a negative utility is assigned to actually hitting someone, and zero utility is assigned to not hitting anyone. Zonal proxemics is more subtle, it relies on the eight proxemic dimensions defined above. It assigns different utilities to the presence of a person in four different zones around an individual which are defined as the *intimate distance*, the *personal distance*, *social distance* and the *public distance* [90]. Gorrini *et al.* [82] studied the proxemics behaviour of groups of pedestrians in interaction and showed that it has negative effects in walking speed for evacuation scenarios. Manenti *et al.* [139] presented an agent-based pedestrian behaviour model that takes into account proxemics and group behaviour. Their model was tested with groups of people and in a simulated environment. A detailed review on proxemics models for robot navigation among humans is proposed in [173].

*b) Physical Models:* These are splitted into three subcategories. The *utility maximization model*, as used in [114], assumes that pedestrian behaves such as to maximize their utility, for example their speed of motion and approach or avoidance of some objects or persons. In the *magnetic force model* proposed by [151], the pedestrian behavior is determined by the equation of motion of the magnetic field. Pedestrians are positive poles and their destinations are negative poles. In the *social force model*, introduced by Helbing [93], each pedestrian has a desired velocity, a target time and a target destination which are affected by social forces such as the interaction with other pedestrians and the effects of the environment. In [133] social forces are described as individual forces (fidelity, constancy) and group forces (attraction, repulsion, coherence). Most of the time, social forces are modeled such that to minimize an energy objective which include terms for individual and group forces.

*c) Cellular-based Models:* These represent a cost model such as Blue and Adler's cellular automata model [23] and used for motion prediction. Cellular Automata (CA) is a discrete, time based modelling formalism on a regular cell grid. It describes the walk of a pedestrian according to rules of a cell occupancy, e.g. a cell can be occupied only if it is free and a pedestrian can have three possible movements: lateral, longitudinal or mitigation of the conflicts. The benefit cost model, developed by Gipps and Marksjo [79], is a discrete and deterministic model where the space is divided into a grid of cells and each agent is described as a particle in a cell. A benefit value, equivalent to the pedestrian utility, is arbitrarily assigned to each cell. In [60] a cellular automata model simulates multi-agent interactions.

*d) Queuing Network Models:* They have been developed for studies of evacuation dynamics [134]. These are evaluated by Monte Carlo simulation methods for discrete events. Each pedestrian is represented as an individual flow entity interacting with other objects, facilities are modeled as a network of arches for openings and of nodes for rooms. In [13], a queuing network model is compared to a social force model for pedestrian crossing movement prediction.

*2) Two Agents' Interaction:* These models are those involving only two agents with mutually influencing behaviours, rather than larger groups of agents. They may be simpler than larger group models but sometimes provide a foundation for extension to larger group models, hence they are here presented first.

*a) Dynamic Graphical Models:* The method in [31] uses POMDPs (Partially Observable Markov Decision Processes) with a time-indexed state space to model interactions and they used the example of an elevator-riding task to test the model. In [179], Rudenko *et al.* proposed a method that uses MDPs with a joint random walk stochastic policy sampling algorithm to predict motion and social forces to model interactions. The model in [120] learns features from observed pedestrian behaviors using a Markov Chain Monte Carlo (MCMC) sampling and performs a Turing test with human participants to validate the human-like behavior of the model. Chen *et al.* [48] used an extended Kalman filter to predict future motions of pedestrians and estimate the time-to-collision range (TTCR) for collision risk level identification.

*b) Gaussian Process Methods:* Kawamoto *et al.* [112] proposed a method to learn pedestrian dynamics with kriging, the most traditional form of Gaussian Processes. Their work can predict pedestrian movement using spatial kriging and spatio-temporal kriging. Social interaction is modeled by spatio-temporal correlation of pedestrian dynamics and correlation is estimated by kriging.

*c) Deep Learning Methods:* Alahi *et al.* [2] predicted pedestrian trajectories in crowded spaces using a social LSTM, a variant of recurrent neural network model that can learn human movement (velocity, acceleration, gait...) taking into account social human motion conventions and predict their future trajectories. This technique is opposed to traditional social forces methods and outperforms most the state-of-art methods on public datasets (ETH and UCY). Long Short-Term Memory (LSTM) can learn and reproduce long sequences, it is a data-driven technique. One LSTM is used for each person and the interaction among people is modeled by a social pooling layer which allows the share of states between neighboring LSTMs. Although group behavior is not modeled, the social LSTM can predict it very well. Similarly to the previous method, Chen *et al.* [49] developed a long-term pedestrian prediction model using RNNs for pedestrian trajectory prediction.

*d) Road Crossing Models:* This section extends the event-activity models from section II-D by adding interaction between pedestrians and vehicles. When microscopic models of pedestrian movement are included in larger-scale traffic simulations together with vehicles, they are typically extended with specific provisions to account for pedestrian's decisions on where and when to initiate road crossing, when this is needed for the pedestrians to reach their goals. Other, so called *gap acceptance* models, have instead described probabilities of pedestrians crossing in a certain gap between vehicles, using generalised linear models, with predictors including both the available gap itself, as well as other factors such as age and gender of the pedestrians, number of pedestrians waiting to cross, and time spent waiting [195], [188].

Markkula *et al.* [140] proposed another type of model for pedestrians road crossing decision, modelled as the result of a number of perceptual decisions concerning the available gap, but also car yielding, explicit communicative signals from the car, and eye contact with the driver. These decisions were described as several interconnected evidence accumulation processes, and it was shown that empirically observed bimodal distributions of pedestrian waiting time were qualitatively reproduced by the model. In [40], Camara *et al.* proposed a heuristic model for pedestrian crossing intention estimation. Their method is based on a distance ratio model that computes the pedestrian crossing probability over time until the curbside. Their results showed that this heuristic model is sufficient for most of the crossing scenarios present in the dataset used and that the remaining scenarios would require higher level models such as game theory.

*e) Other Methods:* Discrete choice models [8], [28] offer a framework to model pedestrian walking along link levels, where their paths are composed of a sequence of straight lines in absence of obstacles. For example, the model in [28] predicts pedestrian behaviour in the presence of other people in shopping street areas.

*3) Group Interaction:* A *group* is here considered to be a collection of more than two pedestrians, but smaller and more cohesive than a crowd. These models are developed primarily for use by non-carriageway autonomous vehicles, such as delivery robots, navigating through crowded pedestrianized areas, needing to cut their way between groups.

*a) Dynamic Graphical Models:* In [19] a real-time pedestrian path prediction is performed in cluttered environments without making any assumption on pedestrian motion or pedestrian density. Pedestrian motion and movements patterns are learnt from 2D trajectories. Bera *et al.* used sparse and noisy trajectories data from indoor and outdoor crowd videos. By combining local movements (microscopic and macroscopic motion models) and global movements (movement flow), the patterns help improve the accuracy of the long-term prediction. An ensemble Kalman filter (EnKF) was used to predict the next state based on current observation and EM algorithm to maximize the likelihood of the state. Pedestrian clusters are computed based on their positions, velocities, inter-pedestrian-distances, orientations etc. Global movement patterns are the past movement and intended velocity of pedestrians. Local movement patterns are obtained by fitting the best motion model to pedestrian clusters and individual motions. In [20], the same authors implemented a tracking algorithm built on top of [19]. Deo *et al.* in [55] uses VGMMs to model pedestrian trajectory using pedestrian origins and destinations. Their model is tested on a dataset of a crowded unsignalized intersection in a university campus. Pellegrini *et al.* [156] introduced a linear trajectory avoidance (LTA) model which has similarities with the social force model. In [157], the same authors extended the LTA model with a stochastic version taking into account group behavior and allows multiple hypotheses about the pedestrian position. Zhou *et al.* [231] proposed a mixture model of dynamic pedestrian-agents (MDA) for pedestrian trajectory prediction in crowds.

*b) Gaussian Process Methods:* Henry *et al.* [94] used inverse reinforcement learning (IRL) to learn human-like navigation behavior in crowds. The model estimates environmental features using Gaussian Processes and extends Maximum Entropy Inverse Reinforcement Learning (MaxEnt IRL) of [232] by assuming that features in the environment are partially observable and dynamic. The proposed approach was developed for mobile robot motion planning, but it could be used for human motion prediction. In [203], Trautman and Krause proposed to solve the freezing robot problem, where a robot motion planner gets stuck and cannot find any proper move to perform, by a model based on Gaussian Processes, a statistical model that is able to estimate crowd interaction.

*c) Deep Learning Methods:* The subsequent models may not explicitly consider interaction, but they learn interaction implicitly through machine learning techniques. The model in [196] implemented a real-time Temporal 3DOF-Pose Long-Short-Term Memory using 3D lidar data from a mobile robot. Shi *et al.* [193] developed a long-term pedestrian trajectory prediction model for crowded environments using LSTM. In [224], Yi et al. proposed a deep neural network model called behavior-CNN that is trained with crowded scenes video data. A pedestrian behavior model is encoded from the previous frames and used as an input for the CNN model to predict their future walking path and destination as well as a predictor for a tracking system. Radwan *et al.* [164] presented an interaction-aware TCNN, a convolutional neural network model that can predict interactive motion of multiple pedestrians in urban areas.

Amirian *et al.* [6] predicted the motion of pedestrians over a few seconds, given a set of observations of their own past motion and of those of the pedestrians sharing the same space, using a Generative Adversarial Network (GAN)-based trajectory sampler. The reason for this choice is that such a method naturally encompasses the uncertainty and the potential multi-modality of the pedestrian steering decision, which is of great importance when using this predictive distribution as a belief in higher level decision-making processes. Lee *et al.* [122] developed *DESIRE* a trajectory prediction framework for multiple interacting agents based on deep neural networks. A conditional variational auto-encoder is used to generate hypothetical future trajectories. An RNN is then used to score and rank those features in an inverse optimal control manner and taking into account the scene context. Gupta *et al.* [85] proposed a socially-aware GAN with RNNs for pedestrian motion sequence prediction in dynamic environments. However, their model assumes that people influence each other uniformly. A detailed analysis and improvement of this GAN method is proposed in [118]. With a similar method, called SoPhie, Sadeghian *et al.* [181] developed a GAN-based trajectory prediction model that focuses on the most important agents for each interacting agent.

*d) Other Methods:* Moussaid *et al.* [148] presented a heuristics-based model to predict pedestrian behavior in crowded environments. Based on the idea that visual information is very important for pedestrians [12], [204], they found that two simple heuristics can model the interaction among people: the desired walking direction and speed of pedestrians are sufficient. Bonneaud and Warren [26] proposed a related type of model, extending the behavioral dynamics model by [72] to goal-seeking and obstacle avoidance in crowds, and found that the model was able to reproduce qualitative crowd phenomena like lane formation. The model in [101] learns behavioral patterns from pedestrian trajectories in a mall. It assumes that a robot can model interactions using social forces and segment pedestrian trajectories into sub-goals to estimate their future positions.

*B. Macroscopic Models*

In macroscopic models, the crowd is modeled as a single ontological object, replacing and simplifying the representation of multiple microscopic pedestrians. The crowd behaves as a continuous fluid with a flow average speed [199].

The first macroscopic models of pedestrians are due to Hughes and Henderson [99]. The fluid dynamic model classi-fies pedestrians into groups which are characterized by average features, their position, speed and intended velocity. In [14], pedestrian flows are modeled in simulations for crowded environments. Crowd modelling has also an established community focused on models for evacuation, as reviewed in [184]. In [4] Ali *et al.* used Lagrangian Particle Dynamics to segment high density crowd flows. This method, based on Lagrangian Coherent Structures (LCS) from fluid dynamics and particle advection, is capable of detecting instabilities in the crowd.

Smooth Particle Hydrodynamics (SPH) is a hybrid of microscopic and macroscopic models. Pedestrians are considered individually, but at each time they are aggregated into a density where each particle is moved according to the macroscopic velocity. Etikyala *et al.* [70] reviewed smooth particle hydrodynamics pedestrian flow models while [225] proposed a generic SPH framework for modeling pedestrian flow.

### C. Discussion

The theory of proxemics has been well studied in psychology and now being more and more used for VR experiments [152] [58] and computer scientists are just beginning to apply it to make more detailed models of the utility of pedestrian's personal space than simply collisions and non-collisions. In general, microscopic models are preferred to macroscopic models, in particular the social force model is very popular for pedestrian interaction modelling, while macroscopic models are more suited for crowd behaviour modeling, especially in the specialised domain of emergency evacuation modeling. Physical models bring interesting results when there are a lot of interactions, e.g. modelling pedestrian movement in cities [177]. Cellular-based models are useful for modelling pedestrians with minimal movement choices and when representing their collisions is not required. Two agents' and group interaction models offer more precise pedestrian models but they require more computational resources, in particular dynamic graphical, Gaussian Process and deep learning models. More computational research is needed in interaction modelling: psychology/human factors studies and theories are more mature, but their results have not yet been quantified to the extent of enabling translation into algorithms for AVs.

## IV. GAME THEORETIC AND SIGNALLING MODELS

### A. Game Theory Interaction Models

The models in section II predict the behavior of a single pedestrian $X$ from the point of view of an external observer $O$ (i.e. the experimenter), when no other pedestrians are present. We call this a first-order model of pedestrian behaviour.

The models in section III all further allow $O$ to also model $X$s own first-order model of another pedestrian $Y$s behaviour, which $X$ can use to plan to avoid $Y$. We call this a second-order model of behaviour.

We could then imagine third and higher order models. For example, $O$ might model $X$'s belief about $Y$'s belief about $X$'s belief about $Y$'s belief, as both agents try to 'out-think' each other during their planning. This would lead to an infinite computational regress.

Game theory provides an alternative and stronger framework which can compute the infinite limit of these higher order models directly, via analytic solutions.

Isaacs [106] introduced vehicle-pedestrian interactions as the famous 'homicidal taxi driver problem which considered the inverse of the modern AV interaction problem: how an AV controller should act in order to *hit* a pedestrian [2]he application to pedestrians was accidental as the taxi scenario was used initially as a declassification technique to publish missile-defence algorithms, requiring control of one missile to hit another. Game theory is in common use in descriptive road user modelling as reviewed in [67], where applications include modelling of lane changes and merging onto motorways, route selection and departure time in congested networks, and socio-economic choices such as purchasing large vehicles or using conventions such as headlight dipping. It has been applied to AV-vehicle interactions in [165] though here only pedestrian models are considered.

The use of game theory for active control of AVs is less common. Descriptive models may be incomplete as active controllers, in particular by allowing for multiple Nash Equilibria to exist without selecting between them. A Nash equilibrium is a set of probabilistic strategies to be played by each of the players, such that no player would change their strategy if they knew the strategies of the other players. It is generally agreed in Game Theory that it is not optimal for players to employ strategies which are not Nash equilibria, though there is still philosophical debate over what strategy is optimal when multiple equilibria exist.

*1) Two Agents' Game Theory Interactions:* Hoogendoorn and Bovy [97] give a purely theoretic construction (left as an exercise to the reader) for a continuous ('differential) game theory solution to pedestrian interactions, based on similar control theory models to those reviewed in Sec. II-A. They also provide an implementation of a second-order truncation of this model which is found to be sufficient to produce flows of pedestrians in crowded environments similar to those observed in some Japanese crossings.

The methods in [141] and [206] predict selection of pedestrian trajectories from a finite set as a higher-order model. For a small set of known origins and destinations, optimal free space trajectories are computed from control theory, and actual trajectories from a video set are compared to them and assigned costs according to their deviations from them. These models assume that the choice of the entire continuous trajectory is drawn from a finite set of previously observed and costed trajectories as a single decision at the start of the interaction and does not model responses to the other agent during the interaction. They are used only as descriptive models rather than as real-time control because they require each pedestrian's final goal location to be known in advance to form the cost matrix – which is only obtainable by looking ahead in the data to see what happened *post hoc*. The authors state that (in the context of AV control), 'few researchers have considered interaction between (pedestrian) objects, thus neglecting that humans give way to each other'. Turnwald *et*

[2]T

*al.* [205] adds an alternative model where one player chooses their trajectory first then the second chooses theirs in response to seeing their initial motion.

Ma *et al.* [135] proposed a long-term game-theoretic prediction of interacting pedestrian trajectories from a single starting image. For each future time in the prediction sequence, fictitious play is used to converge the probabilities of each pedestrian's actions to one (of possibly many) Nash equilibrium. The fictitious play assumes that each pedestrian has a known destination goal, some known visual features (age, gender, initial body heading etc) and a known utility function. The utility function scores vectors of word-state features which contain all of (1) the pedestrian's own future trajectory (which may include control theory style costs); (2) probabilistic beliefs about the other agents' trajectories; (3) the pedestrian's own visual features (age, heading etc); (4) proximity to static obstacles; (5) the pedestrian's distance to their goal. Unusually, the utility functions are learned entirely automatically from video data of actualized trajectories, rather than set by theories. Where theory-like behaviours such as proxemics and social forces are observed in simulations, they arise entirely from this learning process. The functions are assumed to be a weighted linear function of the features and a reinforcement-learning-style model is used to obtain per-state values from the full trajectories during learning. A (deep learning) classifier is used to obtain the visual demographic and heading features from annotated training examples. Performance is degraded when the pedestrian's goal locations are not known and are set to be completely uncertain in the feature vectors.

In [75], Fox *et al.* presented a version of the game-theoretic 'game of chicken' for autonomous vehicle-pedestrian interactions at unsignalized intersections. The obtained discrete model called the 'sequential chicken' model allows two players to choose a set of two speeds: decelerate or continue. A new method to compute Nash equilibria is presented, called 'meta-strategy convergence, used for equilibrium selection. Camara *et al.* [41], [34] evaluated the model [75] by fitting one parameter $\theta$ to controlled laboratory experiments where pedestrians were asked to play sequential chicken. This behavioural parameter $\theta$ was found to be a ratio between the utility of avoiding a collision and the utility of saving time. A summary of the work using the sequential chicken model is provided in [37].

*2) Small Group Game Theory Models:* Vascon *et al.* [208] proposed a game theory model for detecting conversational groups of pedestrians from video data, based on the socio-psychological concept of an $F$-formation and the empirical geometries of these formations. Johora and Müller [109] proposed a three-layer trajectory prediction model composed of a trajectory planner, a force-based (social force) model and a game theoretic decision model. The game theory model is based on Stackelberg games, a sequential leader-follower game where pedestrians have three different possible actions: continue, decelerate and deviate and the car has two possible actions: continue and decelerate. This model is able to handle several interactions at the same time.

*3) Crowd Game Theory Models:* Mesmer *et al.* [143] modelled pedestrians' decision-making and interactions during evacuations with game theory. In [192] a model of pedestrian behavior in an evacuation used game theory and showed that pedestrians get greater benefits by cooperating.

### B. Signalling Interaction Models

Signalling models extend interaction models by allowing both the pedestrian and the AV to model and predict each other's actions of giving and receiving pure information, rather than communicating only through their physical poses.

Nathanael *et al.* [149] has proposed a stratified model of mutual awareness between pedestrians and vehicles including AVs. The actor's awareness is divided into three levels, i.e., (1) unaware of the others, (2) factually aware of the other, or (3) aware and actively attending to the other. When one of the two agents is unaware of the other, the interaction may be as simple as collision avoidance by the one aware, relying only on bodily and kinematic cues. When both agents are aware of each other, the interaction takes the form of mutual coordination through implicit cues, whereas when both agents are attentive to each other (as evidenced through eye contact between human actors), the interaction may involve direct communication through explicit signals, such as gestures, nodding etc. In addition attentiveness, as opposed to mere awareness, designates that any physical action from an attentive agent is a response explicitly addressed to the agent at the focus of attention (i.e. it also has a signalling function).

This line of research raises an epistemological question about signalling-based interaction. Some of the models above involve the concepts not just of an agent (1) knowing that the other agent is there, and (2) acting to show the other agent that they are present; but also higher-order knowing and showing these facts. This includes (3) knowing that the other knows they are there and (4) showing the other that they know that the other knows they are there. But also includes arbitrarily higher orders, such as 'knowing that the other has showed that they know that they know that the other knows and so on. There appears to be a potentially infinite regress here, though intuitively most humans find it difficult to comprehend many more levels than the four mentioned here. But it is difficult to argue for why any cut-off should occur at this or other specific level. Intuitively: when two agents make eye contact, they assume that they both then come to know the infinite stack of such statements about each other.

*1) Signals from Pedestrian to Vehicle:* The need for precise eye contact as opposed to simple head direction or gaze towards the vehicle is controversial. Considering gaze or head orientation towards vehicles, there is evidence that pedestrians who initiate crossings without looking at the oncoming vehicle tend to make drivers more attentive to them by keeping larger safety margins [111]. On the other hand, eye contact between pedestrian and driver tends to increase the probability of the vehicle yielding for pedestrians [87]. The apparent controversy between these findings may be attributed to profound differences in the function of these two behavioural traits. While head orientation towards vehicles typically signifies pedestrian situational awareness to drivers, eye contact most probably signifies driver awareness of the pedestrian to the latter [168].

In addition, eye contact is reported to play a non-trivial role in the social dynamics between the two. Nathanael et al. [149] in a naturalistic study of driver pedestrian interaction reported that pedestrian head turning towards a vehicle was sufficient for drivers to confidently infer pedestrians intent in 52% of interaction cases observed. In retrospective think-aloud sessions of their interaction with pedestrians, drivers mentioned pedestrian active head movement and orientation as an important indication of pedestrian awareness of their vehicle. Mutual eye contact between driver and pedestrian was observed only in 13% of interaction cases, accompanied by explicit signalling in 2% of total cases. This is consistent with recent research [167] that reported head orientation/gaze towards vehicles as the most prominent cues for predicting pedestrian intent. In addition, computational models have shown that head direction is a useful trait for pedestrian path prediction and state of situation awareness such as in [39] which argued that if a pedestrian looks at the vehicle, they are less likely to cross the road.

Matthews *et al.* [142] studied pedestrians' behavior with an autonomous goal car equipped with an Intent Communication System (ICS) based on Decentralized MDP to model the uncertainty associated with pedestrian's behavior. Another important factor to take into account is the poor pedestrian signal settings. It has been proven that signal indication and timing affect significantly pedestrian behavior and their crossing decisions [3] [103] [104]. Pedestrians can have sudden speed change while crossing, and such sudden behavioral changes may not be expected by conflicting vehicles, which may lead to hazardous situations. In [104], Iryo-Asano and Alhajyaseen proposed a discrete choice model and Monte Carlo simulation for generating pedestrian speed profiles at crosswalks. In [105], the same authors modelled pedestrian behaviour after the onset of pedestrian flashing green (PFG) via a Monte Carlo simulation. Their results showed a higher probability of pedestrian stopping at longer crosswalks and a significant difference in pedestrian speeds.

Some early steps have however been taken towards modelling at least some levels of explicit knowing and showing of beliefs about each other via signalling behaviour.

*2) Signals from Vehicle to Pedestrian:* Beyond understanding pedestrians signalling behavior, game theoretic models may also enable the AV to give signals to the pedestrians, creating a higher level information game with both players communicating through both their physical actions and also their signals. The full game theory of such interactions has not yet been worked out, and will form part of a complex socio-technical system [175], but there has been notable activity – especially via company patents – in researching displays and other mechanisms for the signalling itself.

Lundgren et al. [132] showed that the lack of two-way communication between driver and pedestrian may reduce pedestrians' confidence to cross the street and their perceived feeling of safety, when crossing. Lichtenthäler et al. [129] reviewed robot trajectories among humans, including identifying needs for additional gestures or motion information such as gaze to communicate intention, which is relevant for last mile delivery. Researchers are currently conducting studies to better understand exactly which information needs to be transferred when interacting with an AV. Schieben et al. [185] propose the following information to be considered by the design team.

- Information about the vehicle automation status
- Information about next manoeuvres
- Information about perception of environment
- Information about cooperation capabilities

To transfer the relevant information, two means of communications can be used for shaping the communication language of an AV. First, pedestrians might benefit from direct communication through the means of external human machine interfaces (eHMIs) [132], [178]. Secondly, also careful design of vehicle movement can be used to explicitly communicate. Risto et al. introduced the term 'movement gestures and found 'advancing', 'slowing early and 'stopping short as commonly used gestures [175]. Consistent with this, Portouli et al. [160] in the context of driver-driver interaction have shown that 'edging was explicitly used by drivers trying to enter a two-way street as a sign of their intent to inform oncoming cars. Studies of human robot interaction have shown that allowing humans to anticipate robot movements by explicit communication through movements of the robots head raises perceived intelligence of the robot even if it did not succeed completing its intended tasks [197], thus overcoming potential machine error through the means of explicit communication. These studies might suggest similar devices such as head-like and eye-like displays for AVs.

While Clamann et al. [51] found mixed influences of explicit communication through novel eHMI on crossing behavior in dynamic traffic situations and argued that pedestrians will largely rely on legacy behavior and not on eHMIs, Habibovic *et al.* [89] found that traffic participants feel calmer, more in control and safer when an eHMI was present on an AV. Petzoldt, Schleinitz, and Banse [158] found that an eHMI can help to convey the intention of a vehicle to give priority to a pedestrian. They also observed that pedestrians needed more time to understand the intention of a vehicle without eHMI in mixed traffic situations [158]. Communicating the intent and awareness of automated vehicles has been considered in a positive way [137] [138]. Habibovic et al [89], [7] argued that, for safety reasons, communication should never be command-based. The vehicle should communicate solely its intentions.

Communication can be directed or undirected. Pedestrians usually assume that any AVs communication is referring to themselves, hence using eHMIs with multiple pedestrians present has to be carried out in a way that minimizes miscommunication (i.e. either letting all pedestrians pass or not displaying a signal at all). Directed signalling minimizes this risk as other road users do not visually perceive the signal of the eHMI. Dietrich *et al.* [59] found that pedestrians were not able to distinguish whether an undirected light signal was addressed to themselves or other traffic participants.Therefore, AVs should either use directed communication in ambiguous situations involving multiple pedestrians or no communication at all, as pedestrians will base their crossing decision on the approaching vehicles kinematics if no eHMI is present. The color of the visual eHMI stimulus may be of importance [218].

The most common eHMI display types are projection, high resolution displays and direct light. Semantics used include animations, concrete iconography, or text. For instance, Habibovic *et al.* developed a communication concept based on external light signals on the top of the windshield [89]. Using various light animations, the intention of the AV as well as the current driving mode such as 'I'm about to yield', 'I'm resting, and 'I'm about to start are displayed on the LED light bar. Clamann *et al.* [51] empirically examines similar models efficacy for giving signals to pedestrians. Further eHMI concepts include mimicking eye contacts by adding visible 'eyes to AVs – based on the well-known tendency for humans to perceive and design faces in cars– which can communicate detection and awareness of pedestrians through eye contact [44], as well as a virtual driver's mimicking furthermore facial expressions or hand signals. In addition to the pure visual-based communication between AVs and other TPs, some concepts also consider a combination of light and audio signals, as in the Google, Uber concepts and Mercedes-Benz concept car F015.

### C. Discussions

Game Theory has a long history of use in V2V (vehicle to vehicle) interactions in classical transport studies, as microscopic models underlying simulations of traffic flows and infrastructure design. Also multi-robot game theory systems are quite mature in robotics. These two streams have not generally been unified or applied to AV-pedestrian modelling, though this is beginning to emerge as an early research area. Like other sophisticated methods, game theoretic models can be computationally expensive and it remains unclear which of their theoretical solutions will have computationally tractable algorithms.

Signalling models remain a distant research frontier. Physical actuators for eHMI signalling are currently being investigated by car manufacturers and recent years have seen much patent activity in the area. But how to best use them to transmit information is not understood. There are currently no game-theoretic models using knowing and showing with explicit signalling but this would appear to be a fruitful area for future research. Eye contact is a particular form of signalling, but even in high level psychology research there remains an ongoing and lively debate about whether it is relevant or useful. The signalling methods reviewed here are mainly from qualitative studies, some work is still needed to implement their findings in algorithms for AVs.

Most of the eHMI concepts presented here do not yet include detailed user studies and thus there remains a need for thorough evaluation including the behavioral and emotional responses of pedestrians in realistic environments. Different findings might be due to different eHMI concepts, diverse traffic scenarios, as well as different communication strategies. While research is still lacking in full understanding of the effects of eHMI on traffic, a large number of conceptual solutions have been proposed. Their influence on pedestrians, regarding their safety, experience and acceptance remains unclear. Most of these conceptual solutions are proposed by industry and involve some form of visual communication as the visual channel is the currently most used channel of communication in traffic as well as the best suited for communication at larger distances in busy environments.

## V. EXPERIMENTAL RESOURCES

### A. Pedestrian Datasets

Large data sets are important resources for training and testing models at all levels, especially when they are annotated with 'ground truth information by humans. Their use has been common for low-level models such as detection and tracking, though there is currently a shortage of high quality annotated data for the higher-level models such as social interactions.

Major visual pedestrian datasets include the Caltech Pedestrian Benchmark [61], ETH [69], TUD-Brussels [220], Daimler [68], Stanford Drone Dataset [176], UCY Zara pedestrian dataset [125] and INRIA [53]. CityPersons [228] is a large dataset for pedestrian detection. Town Center Dataset [16] is a video dataset composed of 71.5k annotations.

Datasets used for pedestrian re-identification, i.e. having many images of the same people with identifiers include for example CUHK01 [127], CUHK02 [126] and CUHK03 [128], collected at a university campus and composed of thousands of bounding boxes of unique people. DUKEMTMC [174] and DUKEMTMC-reID [230] datasets have been developed in the Duke university campus and are used for tracking and re-identifying multiple people with multi-camera systems. MARKET-1501 [229] dataset provides 35k images of 1500 individuals but also comes with a 500k dataset of non-pedestrian street window distractors for training classifiers. Multi-Object Tracking Benchmark [144] collects diverse datasets and publishes new data. Several releases have already appeared: MOT15, MOT16 and MOT17.

PETA benchmark [54] is a mixture of several public datasets (e.g VIPER, SARC3D, PRID, MIT, I-LID, GRID, CAVIAR4REID, 3DPES), which has been used to recognize pedestrian attributes at far distance. The benchmark has been tested with an SVM method. Social ground truth annotations are much rarer. [38] and [39] collected high quality human annotations of physical and social events during pedestrian-vehicle interactions, including the presence and timings of the agents communicating with each other via eye contact, hand gestures, positions and speeds, and the final 'winners' of interactions which compete for road space during crossings.

Yang *et al.* [223] pointed out that in mixed urban scenarios, intelligent vehicles (IVs) have to cope with a certain number of surrounding pedestrians. Therefore, it is necessary to understand how vehicles and pedestrians interact with each other. They proposed a novel pedestrian trajectory dataset composed of CITR dataset and DUT dataset, so that the pedestrian motion models can be further calibrated and verified, especially when the vehicle's influence on pedestrians plays an important role. In particular, the final trajectories of pedestrians and vehicles were refined by Kalman filters with linear point-mass model and nonlinear bicycle model, respectively, in which $xy$-velocity of pedestrians and longitudinal speed and orientation of vehicles were estimated.

Zhan et al. proposed INTERACTION dataset [227] which contains naturalistic motions of various traffic participants in a variety of highly interactive driving scenarios. Trajectory data was collected using drones and traffic cameras, containing data from multiple countries (USA, China, Germany and Bulgaria). There are four different driving scenarios, with their semantic maps provided: roundabouts, un-signalized intersection, signalized intersection, merging and lane changing. Chang et al. proposed Argoverse [45] containing two datasets and HD maps recorded from a self-driving car. Argoverse 3D Tracking is for 3D object annotations, it contains a collection of 11,052 tracks, and Argoverse Motion Forecasting is a curated collection of 324,557 scenarios, each 5 seconds long, for trajectory prediction. Each scenario contains the 2D, birds-eye-view centroid of each tracked object. ApolloScape dataset [216] was recorded in urban areas in China using various sensors. The dataset contains different road road users (vehicles, pedestrians, bicycles). The ApolloScape LeaderBoard shows the ranking and performance of the models tested on the dataset for different tasks, such as scene parsing, detection/tracking, trajectory prediction, self-localisation. The Intersection Drone (InD) dataset [25] contains naturalistic vehicle trajectories recorded using a drone at four German intersections. It provides the trajectories for thousands of road users and their types (e.g car, pedestrian, bicycle, truck), and can be used for example for road user prediction.

Person detection in off-road agricultural vehicle environments has become popular in recent years. Results from these studies are not well known in transport research but may transfer to on-carriageway and on-pavement AVs as they deal with similar types of pedestrian interactions. The National Robotics Engineering Center (NREC) Agricultural Person Detection Dataset [159] consists of labeled stereo video of people in orange and apple orchards taken from a tractor and a pickup truck, along with vehicle position data. The dataset combines a total of 76k labeled person images and 19k sampled person-free images. Gabriel et al. [76] present a dataset that focuses on action/intention recognition problems for human interactions with small robots in agriculture, including ten actors performing nine gestures and four activities. Stereo camera images, thermal camera images and Lidar point cloud data are recorded on grassland, under varying lighting conditions and distances. Kragh et al. [119] presents a multi-modal dataset for obstacle detection in agriculture containing 2h of raw sensor data from a tractor-mounted sensor system in a grass mowing scenario, including moving humans scattered in the field.

A summary of pedestrian datasets is given in the supplementary material Sect. III Table II.

### B. Vehicle Datasets

To train and test models of pedestrians interacting with vehicles, it is most likely useful to provide similar big data about vehicles as well as about pedestrians. This may include ground truth information on vehicle location and motion, but also high level social annotations to use in studies of interaction with pedestrians. Visual data available includes the Berkeley DeepDrive Video (BDDV) dataset [222], currently the largest vehicle dataset publicly available with 10k hours of driving videos around the world. KITTI dataset [78] provides a one hour video of a vehicle driving in an urban environment. Caesar et al. [32] presented *nuScenes* a dataset for autonomous driving composed of multiple sensor data (RGB, LIDAR, RADAR) from two cities and containing 1k scenes. A summary of vehicle datasets is given in the supplementary material Sec. III Table III.

### C. Pedestrian and Driving Simulators

Three types of relevant simulation research work exist: pedestrian, vehicle, and combined pedestrian-vehicle. Hardware designs and source code for commonly used simulators are often not made public, making it difficult for others researchers to investigate and replicate experiments. So there remains a clear need for more open-source simulators. The open source Godot game and VR engine[3] has recently matured so may soon be used for this purpose. A summary of the simulators is included in the supplementary material Sec. III Table IV.

*a) Pedestrian Simulators:* Pedestrian simulators are VR (Virtual Reality) based environments where pedestrian participants encounter virtual vehicles in order to study pedestrian perception and decision making subject to various oncoming vehicle behaviors [186]. For example, Camara et al. [35], [36] used a HTC Vive VR headset for pedestrians interacting with a game theoretic autonomous vehicle. Results showed that VR is a reliable setup for measuring human behaviour for the development and testing of AV technology. Mahadevan et al. [136] presented OnFoot, a VR pedestrian simulator that studies pedestrian interactions with autonomous vehicles in a mixed traffic environment. The Technical University of Munich also developed a pedestrian simulator [73] composed of a head-mounted display, a motion capture system and a driving simulator software. This setup could be connected to a driving simulator enabling multi-agent studies while extracting the participants gait during the crossing process. The current setup utilizes Unity (with a VIVE HMD) and is sometimes coupled with VIVE Trackers for a virtual self-representation to create an immersive virtual environment enabling fast implementations and evaluation of eHMI concepts [59]. PedSim [80] is a free crowd simulation software.

*b) Vehicle Simulators:* Vehicle simulators are physical platforms where drivers encounter virtual pedestrians (dummies) in order to study driver yielding behaviors in specific interaction scenarios. Simulators such as [150] studied driver-pedestrian interactions in mixed traffic environments using a driving simulator (DriveSafetys DS-600c Research Simulator). JARI-ARV (Augmented Reality Vehicles) [108] is a road running driving simulator and JARI-OVDS (Omnidirectional View Driving Simulator is a driving simulator with 360-degree spherical screen and a rocking device. The University of Iowa [207] has developed a driving simulator. A previous review on driving simulators is presented in [194].

---
[3] www.godotengine.org

*c) Pedestrian-Vehicle Simulators:* Micro or macro simulations model both pedestrian and vehicle behavior. Most of these simulations rely on sets of behavioral rules for both agents. These simulators are primarily used for road design purposes and for policy decisions such as the cellular automata-based simulators proposed in [74] and [131] where vehicle-pedestrian crossing behaviour is studied at crosswalks. Feliciani *et al.* [74] further evaluated the necessity of introducing a new crosswalk and/or switching to a traffic light. Chao *et al.* [46] developed a microscopic-based traffic simulator based on a force model to represent the behaviour and interactions between the road users, and aimed for autonomous vehicle development and testing. Chen *et al.* [47] proposed a simulation platform composed of several behaviour models at crosswalks for vehicle-pedestrian conflicts assessment. Gupta *et al.* [86] developed a simulation model, using Matlab and the open-source SUMO (Simulation of Urban Mobility), for autonomous vehicle-pedestrian negotiations at unmarked intersections, considering different pedestrian behaviours. Commercial products include STEPS [147] software for and Legion [124] simulating pedestrian dynamics. VirtuoCity is an example of physical vehicle-pedestrian simulators. It is composed of a pedestrian simulator, HIKER [182], which is a virtual reality 'CAVE-based' environment for pedestrian behavior analysis, a driving simulator [107] and a truck simulator for driver behaviour understanding. IFSTTAR [100] also possesses a pedestrian simulator and developed a driving simulator for driver behavior analysis and human-machine interactions, an immersive simulator for cars, motorcycles and pedestrians behavior simulation, a driving simulator with human assistive devices and a bicycle simulator.

## VI. Conclusions

Pedestrian sensing, detection and kinematic tracking are now well understood and have mature models as reviewed in Part I [33]. Moving from simple kinematic tracking and prediction of pedestrian motions can however depend on extremely high-level models of the state transition required by tracking and prediction. Going far beyond simple random velocity walk models, the present review has shown that there is much scope here to integrate models of pedestrians as intelligent, goal-based, psychological, active, and interactive agents at several levels.

Unlike the more mature methods reviewed in Part I, this review does not recommend particular software implementations for algorithms at these levels, because they remain active research areas rather than completed and standardizable tools. This review finds that many conceptual issues first need to be cleared, before mathematical interfaces – such as probabilities – can be created to link models at these layers, and only then standardized software development can become a reality. (The only exception to this would be for entirely end-to-end machine learning systems, which are not generally considered to be safe or practical due to their lack of transparency.)

At the level of single pedestrian modelling, there now exist good control theoretic models of optimal walking behaviour from known origin to known destination. Here, pedestrians do not usually walk in straight lines, but optimise gradual turning during walking to move in smooth curves. There has been some recent research success in inferring likely destinations from historical data and partial trajectories.

When interaction with other agents is included, models of pedestrians rapidly become more complex and much less well understood. Suboptimal models include only finite orders of epistemological models of pedestrians beliefs, raising the open question of how to handle higher order beliefs about beliefs. Recent game theory approaches have just begun to find optimal behaviours in these higher-order belief cases but only under various simplifying assumptions.

There has been a general shift away from psychology-informed models, using empirical findings such as demographics predicting behaviours, to purely big-data-driven models which learn aspects of such theories internally as black boxes, usually aiming only to predict the behaviour rather than give theoretical explanations of it.

The role of signalling between pedestrians and vehicles during interactions has been studied qualitatively, but is not yet understood at the algorithmic level. Psychologists and road safety designers have evaluated and commercialised many signalling mechanisms, such as flashing of headlights, use of horns, and custom communication light signals. Finding algorithmic strategies to make optimal use of them, and to process information from receiving signals from others, suitable for real-time AV control, remains an open and important question.

# Supplementary Material: Pedestrian Models for Autonomous Driving Part II: High-Level Models of Human Behavior

Fanta Camara[1,2], Nicola Bellotto[2], Serhan Cosar[3], Florian Weber[4], Dimitris Nathanael[5], Matthias Althoff[6], Jingyuan Wu[7], Johannes Ruenz[7], André Dietrich[8], Gustav Markkula[1], Anna Schieben[9], Fabio Tango[10], Natasha Merat[1] and Charles W. Fox[1,2,11]

## I. QUALITY OF CITATIONS

These linked papers (Part I and II) review over 450 papers from high quality journals and conferences such as *CVPR, ICRA, PAMI, IROS, ITSC, ECCV, IV*. It is common in Computer Science fields including machine vision and machine learning for conferences to be considered higher quality or similar quality to journals, while psychology and sociology fields typically consider journals to be more authoritative. The following figures give some ideas about the quality of the cited papers.

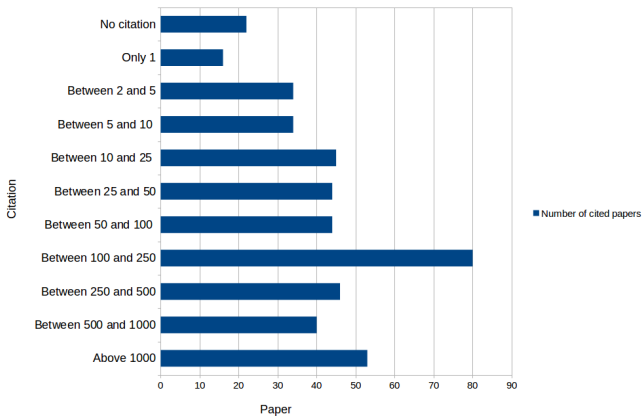

Fig. 1. Number of citations per paper

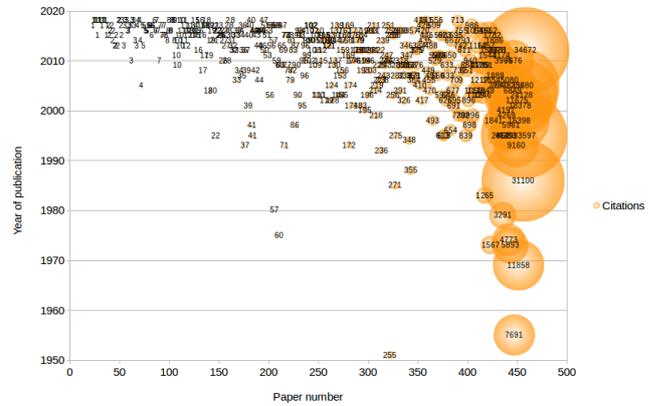

Fig. 2. Number of citations per paper and per year of publication

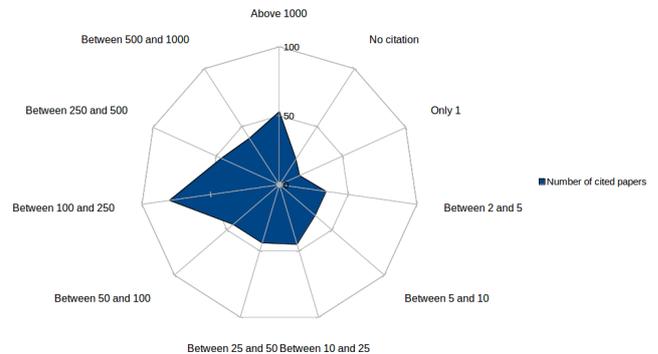

Fig. 3. Number of citations per paper

This project has received funding from EU H2020 interACT (723395).
[1] Institute for Transport Studies (ITS), University of Leeds, UK
[2] Lincoln Centre for Autonomous Systems, University of Lincoln, UK
[3] Institute of Engineering Sciences, De Montfort University, UK
[4] Bayerische Motoren Werke Aktiengesellschaft (BMW), Germany
[5] School of Mechanical Engineering, Nat. Tech. University of Athens
[6] Department of Computer Science, Technische Universität München
[7] Robert Bosch GmbH, Germany
[8] Chair of Ergonomics, Technische Universität München (TUM), Germany
[9] DLR (German Aerospace Center), Germany
[10] Centro Ricerche Fiat (CRF), Italy
[11] Ibex Automation Ltd., UK
Manuscript received 2019-03-11; Revisions: 2019-10-21, 2020-03-26. Accepted: 09-04-2020.

## II. PEDESTRIAN TRAJECTORY AND INTERACTION MODELS

TABLE I: Summary of pedestrian trajectory prediction and interaction models

| Study/Paper | Input/Evaluation | Method | Trajectory Prediction Models | SAE Level |
|---|---|---|---|---|
| Hoogendoorn et al. [1] | Simulated trajectories | Optimal control theory | Unobstructed walking paths | Level [4,5] |
| Antonini et al. [2] | Pedestrian movements (video sequence) | Discrete choice model | Unobstructed walking paths Two agents' interaction | Level 3 |
| Borgers et al. [3] | Pedestrian movements (manual) | Discrete choice | Unobstructed walking paths | Level 3 |
| Puydupin-jamin et al. [4] | Pedestrian trajectories dataset Arechavaleta et al. [5] | Unicycle model with inverse optimal control | Unobstructed walking paths | Level 3 |
| Habibi et al. [6] | Pedestrian trajectories dataset from two intersections [6] | Gaussian Process with a Transferable ANSC algorithm | Route prediction around obstacles Gaussian Process methods | Level [3,4] |
| Kitani et al. [7] | 92 videos (80% for training) | Inverse reinforcement learning with inverse optimal control theory | Uncertain destination models Dynamic graphical models | Level [3,4] |
| Ziebart et al. [8] | Predicted trajectories used in an incremental motion planner | Maximum entropy with inverse optimal control | Uncertain destination models Dynamic graphical models | Level [3,4] |
| Vasquez et al. [9] | 14 pedestrian trajectories dataset [7] | Markov decision process (MDP) with a Fast Marching Method (FMM) | Uncertain destination models Dynamic graphical models | Level 4 |
| Gockley et al. [10] Topp et al. [11] | Laser data | Direction-following and Path-following with Curvature velocity method | Route prediction around obstacles | Level 3 |
| Bennewitz et al. [12], [13] | Laser range data from Pioneer robots | Clustering with Expectation Maximization (EM) algorithm | Uncertain destination models Dynamic graphical models | Level 3 |
| Wu et al. [14] | Pedestrian trajectories from Rutesheim dataset | Markov chains with an heuristic method | Uncertain destination models Dynamic graphical models | Level 3 |
| Karasev et al. [15] | Pedestrian dataset [15] | Markov decision process (MDP) with Rao-Blackwellized filter | Uncertain destination models Dynamic graphical models | Level [4,5] |
| Bai et al. [16] | Autonomous golf car | Partially observable Markov decision process (POMDP) | Uncertain destination models Dynamic graphical models | Level [4,5] |
| Rehder et al. [17] | Pedestrian trajectories (stereo video dataset) | Inverse reinforcement learning | Uncertain destination models Deep learning methods | Level [4, 5] |
| Garzon et al. [18] | Simulations and real-world data | Fast Marching Method (FMM) and A Star (A*) algorithm | Uncertain destination models | Level 4 |
| Kooij et al. [19] | Pedestrian trajectories dataset [19] | Dynamic Bayesian network (DBN) with a switching linear dynamic system (SLDS) | Event/activity models Dynamic graphical methods | Level [4,5] |
| Schulz et al. [20] | Pedestrian dataset [21] | Interacting Multiple Model (IMM) with a latent-dynamic conditional random field (LDCRF) | Event/activity models Dynamic graphical methods | Level [4,5] |
| Dondrup et al. [22] | Laser and RGB-D data | Qualitative Spatial Relations (QSR) with Hidden Markov model (HMM) | Event/activity models Dynamic graphical methods | Level [4,5] |
| Bonnin et al. [23] | Pedestrian datase [23] | Inner-city model with 'Context Model Tree' approach | Event/activity models | Level [4,5] |
| Borgers et al. [24] | Pedestrian dataset from the city of Maastricht | Discrete Choice Model | Event/activity models | Level 3 |
| Camara et al. [25] [26] | Pedestrian-vehicle interactions dataset [25] | Regression models Filtration analysis | Event/activity models | Level [4,5] |
| Völz et al. [27] | LIDAR pedestrian trajectories | Support vector machine (SVM) | Event/activity models | Level 3 |
| Duckworth et al. [28] [29] | Pedestrian dataset from a mobile robot [29] | Qualitative Spatial Analysis (QSR) with a graph representation | Event/activity models | Level [4,5] |
| Mögelmose et al. [30] | Pedestrian trajectories from a monocular camera | Particle filter | Route prediction around obstacles Dynamic graphical methods | [Level 3,4] |
| Schneider et al. [21] | Pedestrian dataset [21] | Extended Kalman filter (EKF) and Interacting Multiple Model (IMM) | Event/activity models Dynamic graphical methods | Level [3,4] |
| Quintero et al. [31], [32] | CMU Dataset with 129 video sequences | Balanced Gaussian process dynamical models (B-GPDMs) and Naive Bayesian classifiers | Event/activity models Dynamic graphical methods | Level[3,4,5] |
| Goldhammer et al. [33] | Camera data | Multilayer perceptron (MLP) with polynomial least square approximation | Uncertain destination models Deep learning methods | Level [4,5] |
| Kruse et al. [34] | Camera data | Statistical analysis | Route prediction around obstacles | Level 3 |
| Cosgun et al. [35] | Real robot | Motion planning with curvature velocity method | Uncertain destination models | Level [3,4] |
| Koschi et al. [36] | Real world data from a moving vehicle | Set-based method Reachability analysis | Uncertain destination models | Level [4,5] |
| Bock et al. [37] | Dataset in [37] | LSTM | Event/activity models Deep learning methods | level 5 |
| Hug et al. [38] | Synthetic test conditions | LSTM with a mixture density output layer (LSTM-MDL) model and particle filter method | Uncertain destination models Deep learning methods | Level [4,5] |
| Cheng et al. [39] | Pedestrian datasets: ETH [40] and UCY [41] | Social-Grid LSTM based on RNN architecture | Uncertain destination models Deep learning methods | Level 5 |
| Bhattacharyya et al. [42] | CityScapes dataset [43] | Two-stream recurrent neural network (RNN) | Uncertain destination models Deep learning methods | Level [4,5] |
| Broz et al. [44] | Simulated data | Time-state aggregated partially observable Markov model (POMDP) | Two agents' interaction | Level 4 |
| Rudenko et al. [45] | Simulated and real data | Markov decision process (MDP) with a joint random walk stochastic policy sampling | Two agents' interaction Graphic dynamical models | Level 5 |

IEEE TRANSACTIONS ON INTELLIGENT TRANSPORTATION SYSTEMS 3TABLE I: Summary of pedestrian trajectory prediction and interaction models

| Study/Paper | Input/Evaluation | Method | Trajectory Prediction Models | SAE Level |
|---|---|---|---|---|
| Kretzschmar et al. [46] | Turing test with human participants | Markov chain Monte Carlo (MCMC) sampling | Two agents' interaction<br>Graphic dynamical models | Level [4,5] |
| Kawamoto et al. [47] | Pedestrian datasets: ETH [40] and [48] | Kriging (Gaussian process) model | Two agents' interaction<br>Gaussian Process methods | Level [3,4] |
| Alahi et al. [49] | Pedestrian datasets: ETH [40] and UCY [41] | Social LSTM | Two agents' interaction<br>Deep learning methods | Level [4,5] |
| Hoogendoorn et al. [50] | Simulations | Optimal control theory | Two agents' interaction | Level [4,5] |
| Ikeda et al. [51] | Shopping mall data | Social force and sub-goal concept | Two agents' interaction | Level [3,4] |
| Chen et al. [52] | Experimental vehicle ALSVIN | Extended Kalman filter (EKF) | Two agents' interaction<br>Graphic dynamical models | Level [3,4] |
| Bera et al. [53] [54] | Indoor and outdoor crowded videos | Ensemble Kalman filter (EnKF) | Group interaction<br>Graphic dynamical models | Level [4,5] |
| Deo et al. [55] [56] | Crowded unsignalized intersection dataset | Variational Gaussian mixture models (VGMM) | Group interaction<br>Graphic dynamical models | Level 5 |
| Pellegrini et al. [40] [57] | Pedestrian dataset with birds-eye view images [40] | Linear Trajectory avoidance model (LTA) | Small Group interaction<br>Graphic dynamical models | Level [4,5] |
| Sun et al. [58] | L-CAS Pedestrian dataset [58] | Temporal 3DOF-pose LSTM (T-pose LSTM) | Group interaction<br>Deep learning methods | Level [4,5] |
| Yi et al. [59] | Crowded scenes video data | Behaviour convolutional neural network (CNN) | Group interaction<br>Deep learning methods | Level [4,4] |
| Radwan et al. [60] | 6 public datasets comprising ETH, UCY, L-CAS | Interaction-aware trajectory convolutional neural network (IA-TCNN) | Group interaction<br>Deep learning methods | Level [4,5] |
| Moussaid et al. [61] | Pedestrian trajectories [61] | Heuristic model | Group interaction | Level 5 |
| Turner et al. [62] | Simulations | Exosomatic visual architecture | Group interaction | Level [4,5] |
| Vasishta et al. [63] | Real world scenes dataset [63] | Natural vision model | Group interaction | Level [4,5] |
| Zhou et al. [64] | Pedestrian dataset from New York Central station | Mixture model of dynamic pedestrian-agents (MDA) | Group interaction<br>Graphic dynamical models | Level [4,5] |
| Shi et al. [65] | 2D laser sensor dataset | LSTM | Group interaction<br>Deep learning | Level [4,5] |
| Amirian et al. [66] | Synthetic dataset | GAN based method with hand-desgined interaction features | Group interaction<br>Deep learning | Level [4,5] |
| Lee et al. [67] | KITTI [68][69] and Stanford Drone Dataset [70] | GAN and RNNs | Group interaction<br>Deep learning | Level [4,5] |
| Gupta et al. [71] | ETH and UCY datasets | GAN and RNNs | Group interaction<br>Deep learning | Level [4,5] |
| Sadeghian et al. [72] | ETH and UCY | GAN | Group interaction<br>Deep learning | Level [4,5] |
| Henry et al. [73] | Crowd flow simulator | Inverse reinforcement learning (IRL) and Gaussian process (GP) | Crowd behaviour models | Level 5 |
| Trautman et al. [74] | Pedestrian dataset: ETH [40] | Gaussian process (GP) | Group interaction<br>Gaussian Process methods | Level 5 |
| Ali et al. [75] | Video from Google videos and National Geographic documentary | Lagrangian particle dynamics model | Macroscopic models<br>Crowd behaviour models | Level 5 |
| Mehran et al. [76] | Dataset of escape panic scenarios and web videos | Particle advection with social force model | Macroscopic models<br>Crowd behaviour models | Level 5 |
| Ma et al. [77] | UCY Zara Dataset, the Town Centre Dataset and the LIDAR Trajectory Dataset. | Fictitious game and reinforcement learning | Game theoretic models<br>Two agents' interaction | Level 5 |
| Isaacs [78] | / | Homicidal taxi driver problem | Game theoretic models<br>Two agents' interaction | Level 5 |
| Turnwald et al. [79] [80] [81] | Pedestrian trajectories from motion capture system [79] | Finite set of single-shot games | Game theoretic models<br>Two agents' interaction | Level 5 |
| Fox et al. [82] [83] | Simulations and dataset in [25] | Game of Chicken | Game theoretic models<br>Two agents' interaction | Level 5 |
| Vascon et al. [84] | Public datasets | Game theory | Game theoretic models<br>Small group interaction | Level 5 |
| Johora et al. [85] | Simulations | Stackelberg games | Game theoretic models<br>Small group interaction | Level 5 |
| Mesmer et al. [86] | Experiments | Game theory with velocity vector | Game theoretic models<br>Crowd interaction | Level 5 |
| Shi et al. [87] | Experiments | Modified lattice model | Game theoretic models<br>Crowd interaction | Level 5 |
| Dimitris et al. [88] | Video data | Two-dimensional classification model | Signalling models | Level 5 |
| Katz et al. [89] | Controlled experiment | Statistical analysis | Signalling models | Level 5 |
| Guéguen et al. [90] | Controlled experiment | Statistical analysis | Signalling models | Level 5 |

## III. DATASETS

TABLE II: Summary of pedestrian datasets

| Dataset | Data type | Viewpoint | Applications | Quantity of data |
|---|---|---|---|---|
| The Caltech Pedestrian Benchmark [91] | Urban video data (Resolution: 640x480) | Moving car | Detection, Tracking, Trajectory Prediction | 10 hours of 30Hz video with 250,000 annotated frames, 350,000 labeled bounding boxes and 2300 unique pedestrians |
| ETHZ Benchmark [92] | Urban video data using a stereo pair of cameras (Resolution: 640x480) | Children's stroller | Detection, tracking, Trajectory prediction | 2,293 frames with 10958 annotations |
| TUD-Brussels [93] | Image pairs | Hand-held camera and Moving car | Detection | Training set: 1,092 positive image pairs (resolution: 720x576) with 1,776 annotations and 192 negative image pairs (resolution: 720x576). Additional 26 image pairs with 183 annotations. Test set: 508 image pairs (resolution: 640x480) with 1,326 annotations |
| Daimler Benchmark [94] | Grayscale camera images | Moving car | Detection | Training set: 15,660 positive samples (resolution: 72 pixels height) and 6,744 negative samples. Test set: 21,790 images with 56,492 annotations including 259 trajectories of fully visible pedestrians |
| INRIA Pedestrian Dataset [95] | Camera images | Any | Detection | 1805 images (resolution: 64x128) |
| CityPersons [96] | Camera images | Moving car | Detection | 5k images with 35k bounding boxes of pedestrians and 20k unique persons |
| Edinburgh Informatics Forum pedestrians overhead dataset [97] | Pedestrian Trajectories | Surveillance camera | Trajectory prediction | Over 92k pedestrian trajectories |
| ETHZ BIWI Walking Pedestrian dataset [40] | Video | Bird-eye view | Detection, Tracking and Trajectory prediction | 650 tracks over 25 minutes |
| UCY Zara pedestrian dataset [41] | Synthesized crowd data | Bird-eye view | Tracking and Trajectory Prediction | 1 video 2-min long with 5-6 persons per frame. 1 video with 40 persons per frame Trajectories |
| Town Center Dataset [48] | Video data | Bird-eye view | Detection, Tracking and Trajectory Prediction | Video (resolution: 1920x1080) with 71500 annotations |
| MARKET-1501 [98] | Camera images | Moving car | Detection and Re-identification | 32k bounding boxes with 1,501 individuals and 500k non-pedestrian (street windows) |
| VIPER Benchmark [99] | Video data | Moving car | Optical flow, semantic instance segmentation, object detection and tracking, object-level 3D scene layout, visual odometry | 250k video frames |
| CUHK01 [100] | Camera images | Surveillance camera | Detection and Re-Identification | 971 persons with 2 camera views |
| CUHK02 [101] | Camera images | Surveillance camera | Detection and Re-Identification | 1,816 unique persons with 5 pair of camera views |
| CUHK03 [102] | Camera images | Surveillance camera | Detection and Re-Identification | 13k images with 1,360 pedestrians |
| DUKEMTMC dataset [103] | Video and trajectories | Surveillance camera | Detection, Tracking, Trajectory Prediction, Re-Identification | 6,791 trajectories for 2,834 unique persons over 85 minutes video per camera (8 cameras in total) (resolution: 1080p) |
| DUKEMTMC-reID dataset [104] | Video and bounding boxes | Surveillance camera | Detection and Re-Identification | Over 36k bounding boxes of 1,812 unique individuals |
| MOTChallenge Benchmark [105] [106] | Video | Any | Detection, Tracking, Re-Identification, Trajectory Prediction | Composed of parts of other datasets and new data (videos, bounding boxes) website: motchallenge.net/ |
| Daimler Pedestrian Benchmark [19] | Annotations | Moving car | Trajectory prediction [crossing, stopping] | 58 annotated pedestrian-vehicle interactions data |
| PETA dataset [107] | Images | Any | Detection and Recognition | 19,000 images with 8,705 persons |
| L-CAS 3DOF Pedestrian Trajectory Prediction Dataset [58] | Pedestrian trajectories | Mobile robot | Trajectory prediction | 50 trajectories |
| L-CAS 3d-point-cloud-people-dataset [108] | 3D LiDAR point clouds | Mobile robot | Pedestrian detection and tracking | 5,492 annotated frames with 6,140 unique persons and 3,054 groups of people |
| IAS-Lab People Tracking dataset [109] | RGB-D video sequences + ground truth given by a motion capture system | Mobile Pioneer P3AT robot | People detection and tracking | 4,671 frames with 12,272 persons |
| Porch experiment dataset [5] | Motion capture system | Any | Trajectory prediction | 1,500 person trajectories |
| UCLA Pedestrian dataset [15] | Video data | Moving car | Trajectory Prediction | 17 annotated video sequences, ranging from 30 to 900 frames, and containing 67 pedestrian trajectories |
| LIDAR Trajectory dataset [77] | Lidar data | Top view | Trajectory Prediction | 20 interacting person trajectories |
| Joint Attention in Autonomous driving (JAAD) [110] | Videos and annotations | Moving car | Detection, Tracking, Trajectory Prediction | 346 video clips with annotations extracted from 240 hours of driving videos |
| MoCap database [111] | Motion capture system | Any | Detection, Recognition | 500k frames with persons in many different poses |
| The Multi-Person PoseTrack Dataset [112] | Video data | Any | Detection, Recognition, Tracking | 60 videos with 16k annotated persons with different poses |
| CMU dataset [7] | Video data | Any | Detection, Tracking, Trajectory Prediction | 92 videos |
| Pedestrian-Vehicle Interactions dataset [25] [26] | Annotations | Human observers | Trajectory Prediction (crossing, stopping) | 204 annotated pedestrian-vehicle interactions at an unsignalized intersection |
| CITR adn DUT datasets [113] | Trajectories | Top view | Trajectory prediction and interaction | over 2k |
| NERC Agricultural Person Detection Dataset [114] | stereo video | Mobile platforms | Detection and Tracking | 76k labelled person images and 19k person-free images |

TABLE II: Summary of pedestrian datasets

| Dataset | Data type | Viewpoint | Applications | Quantity of data |
|---|---|---|---|---|
| Action Recognition Dataset [115] | Stereo camera and thermal images + Lidar point clouds | Mobile robot | Action and gesture recognition | 10 actors performing 9 gestures and 4 acivities |

TABLE III: Summary of vehicle datasets

| Dataset | Data type | Applications | Quantity of data |
|---|---|---|---|
| Berkeley DeepDrive Video dataset (BDDV) [116] | Video and GPS/IMU data | Detection, Tracking, Identification | 10k hours of driving videos around the world |
| EPFL multiview car database [117] | Images | Detection, Identification | 2k images with 20 different car models |
| KITTI dataset [68] [69] | Video data | Detection, Tracking, Identification, Localisation | About 1 hour in one city in daytime |
| Cityscape [43] | Video data | Detection, Tracking, Identification | About 100 hours videos in multiple cities in daytime |
| Commai.ai [118] | Video data | Detection, Tracking, Identification | 7.3 hours videos in highway during daytime and night |
| The Oxford RobotCar Dataset [119] | Video data | Detection, Tracking, Identification | 214 hours videos in Oxford in daytime |
| Princeton TORCS DeepDriving [120] | Synthetic video data | Detection, Tracking, Identification | 13.5 hours videos in highways |
| Honda Research Institute Driving Dataset (HDD) [121] | Video data | Detection, Tracking, Identification | 104 hours videos in one city |
| Udacity [122] | Video data | Detection, Tracking, Identification | 8 hours of videos |
| nuScenes [123] | RGB, LIDAR and RADAR data | Scene understanding | 1k different scenes form 2 cities |
| Fieldsafe dataset [124] | Multiple sensors | Obstacle detection | 2h raw sensor data from a mobile platform |

TABLE IV: Summary of pedestrian and vehicle simulators

| Simulator | Type | Applications |
|---|---|---|
| Technical University of Munich (TUM) | <ul><li>Pedestrian simulator: Head-mounted display with a motion capture system</li><li>Driving Simulator software</li></ul> | <ul><li>Pedestrian behaviour understanding</li><li>Driver behaviour analysis</li></ul> |
| Institute for Transport Studies (ITS), University of Leeds | <ul><li>HIKER Lab : pedestrian simulator</li><li>Driving Simulator</li><li>Truck Simulator</li></ul> | <ul><li>Pedestrian behaviour understanding</li><li>Pedestrian interaction with the environment</li><li>Driver behaviour understanding</li></ul> |
| Japan Automobile Research Institute (JARI) | <ul><li>JARI-ARV (Augmented Reality Vehicle)</li><li>JARI-OVDS (Omni Directional View Driving Simulator)</li></ul> | <ul><li>Road running driving simulator</li><li>Driving simulator with 360-degree spherical screen and rocking device</li></ul> |
| French Institute of Science and Technology for Transport, Development and Networks (IFSTTAR) | <ul><li>Driving Simulator</li><li>Immersive Simulator</li><li>Driving Simulator with human assistive devices</li><li>Bicycle Simulator</li></ul> | <ul><li>Driver behaviour analysis</li><li>Road user behaviour understanding</li></ul> |
| University of Iowa | Driving Simulator | Driver behaviour understanding |
| Pedsim [125] | Synthetic simulator | Crowd behaviour understanding |
| OnFoot [126] | VR pedestrian simulator | Pedestrian behaviour understanding |
| Vehicle-Pedestrian simulators [127] [128] | Cellular automata models | Vehicle-pedestrian interactions |
| Mcroscopic traffic simulator [129] | Force model | Traffic and road user behaviour understanding |
| AV-Pedestrian negotiations simulator [130] | Different pedestrian behaviour models | Pedestrian behaviour understanding for AVs |